\documentclass[twocolumn]{article}
\usepackage{bagrow}
\usepackage{bm}
\usepackage{booktabs}
\usepackage{multirow}
\usepackage{geometry}
\usepackage{enumitem}

\usepackage{xspace}
\newcommand{\rmse}{\mathit{RMSE}\xspace}

\title{Multi-Exit Kolmogorov--Arnold Networks: enhancing accuracy and parsimony}

\author[1,2,*]{James Bagrow}
\author[3,2]{Josh Bongard}
\affil[1]{Mathematics \& Statistics, University of Vermont, Burlington, VT, United States }
\affil[2]{Vermont Complex Systems Center, University of Vermont, Burlington, VT, United States}
\affil[3]{Computer Science, University of Vermont, Burlington, VT, United States }

\affil[*]{\corrauthinfo{james.bagrow@uvm.edu}{bagrow.com}
}

\date{\today}

\begin{document}

\singlespacing

\maketitle

\begin{abstract}
Kolmogorov--Arnold Networks (KANs) uniquely combine high accuracy with interpretability, making them valuable for scientific modeling.
However, it is unclear \emph{a priori} how deep a network needs to be for any given task, and %
deeper KANs can be difficult to optimize and interpret.
Here we introduce multi-exit KANs, where each layer includes its own prediction branch, enabling the network to make accurate predictions at multiple depths simultaneously. 
This architecture provides deep supervision that improves training while discovering the right level of model complexity for each task. 
Multi-exit KANs consistently outperform standard, single-exit versions on synthetic functions, dynamical systems, and real-world datasets. 
Remarkably, the best predictions often come from earlier, simpler exits, revealing that these networks naturally identify smaller, more parsimonious and interpretable models without sacrificing accuracy. 
To automate this discovery, we develop a differentiable ``learning-to-exit'' algorithm that balances contributions from exits during training.
Our approach offers scientists a practical way to achieve both high performance and interpretability, addressing a fundamental challenge in machine learning for scientific discovery.
\end{abstract}


\section{Introduction}

Machine learning has become indispensable for scientific discovery, enabling researchers to uncover complex patterns in data that traditional analytical methods cannot readily capture~\cite{carleo2019machine,roscher2020explainable,xu2021artificial,wang2023scientific}.
Neural networks and related techniques excel at nonlinear regression and data-driven modeling of dynamical systems---fundamental challenges spanning physics, biology, chemistry, and engineering~\cite{carleo2019machine,karniadakis2021physics,wang2023scientific}. 
From climate modeling~\cite{kashinath2021physics,doi:10.1126/science.adi2336} to protein folding prediction~\cite{jumper2021highly}, these data-driven approaches can learn sophisticated functional relationships directly from observations, opening new pathways to understanding natural phenomena where first-principles models are unavailable or computationally intractable~\cite{MONTANS2019845,roscher2020explainable,bradley2022perspectives,wang2023scientific}.

Despite these advances, scientific applications face a fundamental challenge that is less pressing in many other machine learning domains: the need to achieve simultaneously both high predictive accuracy and model interpretability~\cite{roscher2020explainable,bell2022s}. 
While many commercial applications prioritize accuracy above all else, scientific modeling demands that researchers understand not just what the model predicts, but how and why it makes those predictions~\cite{roscher2020explainable,wang2023scientific}. 
Interpretability is essential for determining whether models capture genuine physical relationships rather than spurious correlations, for gaining scientific insight into the underlying phenomena, and for building the trust necessary to guide experimental design or inform policy decisions~\cite{ferrario2022explainability,bell2022s,van2023ai}. 
Unfortunately, accuracy and interpretability are in tension: the most accurate machine learning models---typically deep neural networks with millions or billions of parameters---are often the least interpretable, functioning as ``black boxes'' that provide little insight into the mechanisms driving their predictions~\cite{castelvecchi2016can,wang2023scientific}. 
This accuracy--interpretability trade-off remains a critical bottleneck for the adoption of machine learning in scientific discovery, where understanding the underlying relationships is often as important as making accurate predictions~\cite{roscher2020explainable,wang2023scientific}.

Kolmogorov--Arnold Networks (KANs) have recently emerged as a promising solution to this accuracy-interpretability dilemma, representing one of the rare neural architectures that can achieve both high predictive performance and meaningful interpretability~\cite{liu2025kan,liu2024kan2.0,toscano2025pinns}. 
Motivated by the Kolmogorov--Arnold Representation Theorem, which shows that multivariate functions can be represented as compositions of univariate functions, KANs provide a divide-and-conquer approach to high-dimensional problems by breaking them down into manageable univariate components that can be learned directly from data~\cite{liu2025kan}. 
This enables KANs to discover and represent complex functional relationships while maintaining the ability to visualize and interpret each learned univariate function individually. 
Results have demonstrated that KANs can achieve competitive accuracy with traditional deep networks on regression tasks and dynamical systems modeling while providing insights into the learned functional forms~\cite{liu2025kan,liu2024kan2.0,koenig2024kan,PhysRevResearch.7.023037}. 
This combination of accuracy and interpretability makes KANs well-suited for scientific applications where understanding the underlying mathematical relationships is as crucial as predictive performance.

Despite these promising qualities, KANs face challenges that can limit their effectiveness in scientific applications. 
Training KANs presents optimization difficulties, as the learnable univariate functions must be carefully parameterized and refined, with deeper networks often proving particularly challenging to optimize~\cite{liu2025kan}. 
The architecture search problem---determining the appropriate number of layers and widths---also remains significant, as practitioners must balance expressiveness against parsimony while avoiding overfitting~\cite{elsken2019neural}. 
Seeking smaller, more parsimonious models without sacrificing accuracy is crucial because overly deep KANs lose interpretability as compositions of many univariate functions become difficult to understand, while smaller KANs better retain the interpretability that makes them valuable for scientific modeling~\cite{liu2025kan}. 
These challenges suggest that architectural innovations are needed to better realize KANs' potential for scientific discovery.

\begin{figure*}[t!]
    \centering
    \includegraphics[width=0.85\linewidth]{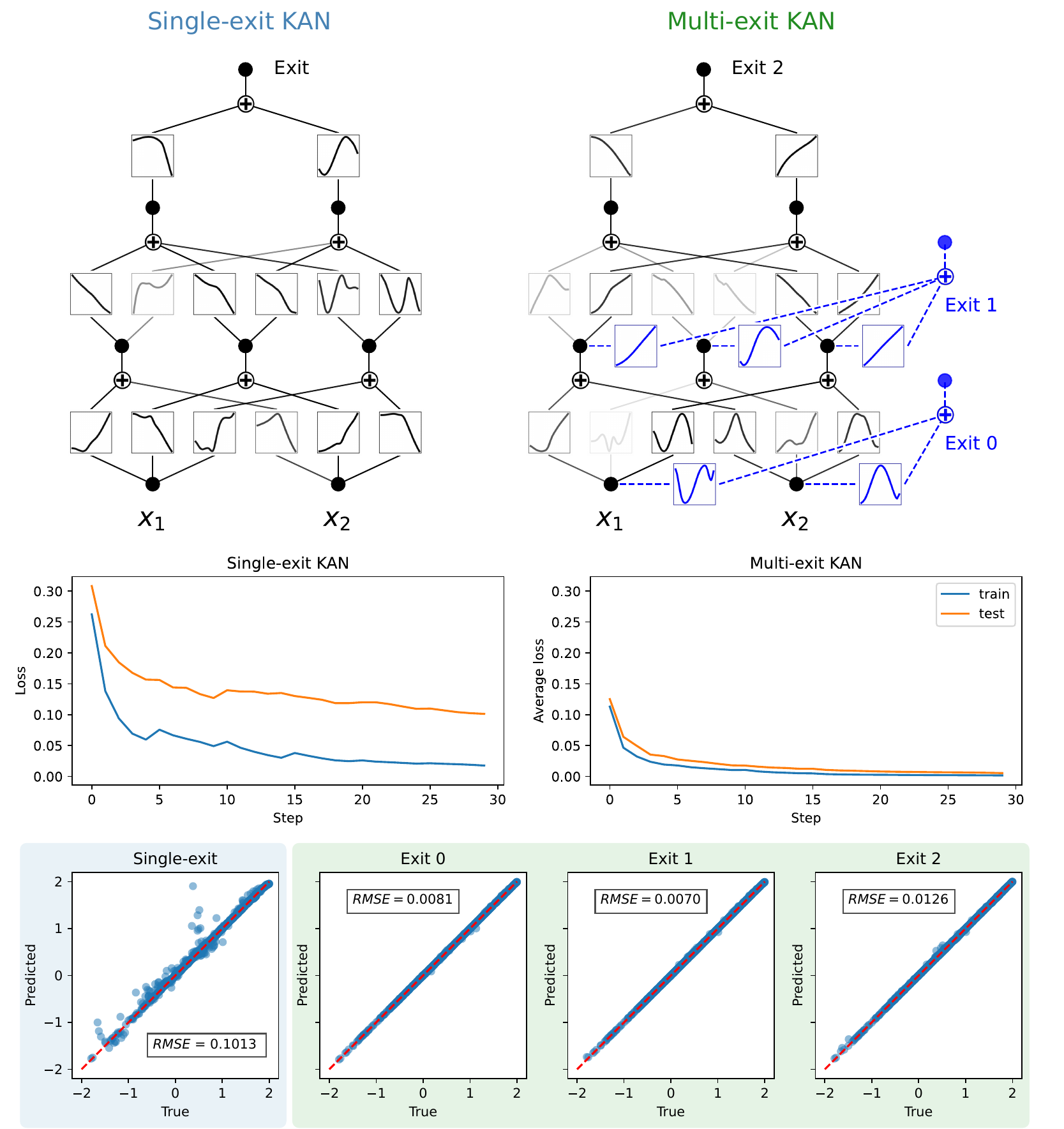}
    \caption{Enhancing accuracy and interpretability of Kolmogorov--Arnold networks (KANs) with multiple exits, here illustrated on the toy problem $y = \sin\left(x_1\right) + \cos\left(x_2\right)$.}
    \label{fig:overview}
\end{figure*}

In this paper, we introduce \textit{multi-exit architectures}~\cite{lee2015deeply,branchyNet} into KANs as a novel approach to address these challenges while preserving KANs' interpretability advantages. 
Multi-exit networks, originally developed for deep networks to enable adaptive inference and computational efficiency, augment networks with additional prediction branches at intermediate layers, allowing models to make predictions at multiple depths~\cite{branchyNet, conditionalDeepLearning, scardapane2020should}. 
When applied to KANs, this approach offers a novel way to tackle the architecture search challenge by enabling a single network to effectively explore multiple levels of complexity simultaneously~\cite{scardapane2020should}. 
Multi-exit KANs can help identify appropriate levels of model complexity for a given task: if early exits perform well, the network has found a parsimonious and more interpretable model that maintains accuracy, while deeper exits remain available when additional expressiveness is needed (Fig.~\ref{fig:overview}). 
Furthermore, the multi-exit approach enables deep supervision~\cite{he2016deep} during training, where gradients flow directly to earlier layers through the exit branches, potentially improving the optimization of deeper KANs.

Experiments demonstrate the effectiveness of multi-exit KANs across diverse scientific modeling tasks, showing consistent improvements in both accuracy and parsimony compared to traditional single-exit KANs. 
Our key contributions include: 
    (1) the first application of multi-exit architectures to KANs, with a joint training framework that enables deep supervision; 
    (2) empirical validation across regression problems, dynamical systems modeling, continual learning scenarios, and real-world datasets; 
    (3) evidence that multi-exit KANs often achieve better performance at earlier exits, indicating more parsimonious models without sacrificing accuracy; and 
    (4) a ``learning-to-exit'' algorithm that addresses the choice of extra hyperparameters when using multi-exits. 
Additionally, we provide insights into why multi-exit architectures are particularly well-suited for KANs and discuss the mechanisms underlying their improved performance. 
These results suggest that multi-exit architectures represent a valuable enhancement to KANs for scientific applications, offering a principled approach to balancing accuracy and interpretability.

The rest of this paper is organized as follows. 
Section~\ref{sec:background} provides background on Kolmogorov--Arnold networks and multi-exit and early-exit neural architectures. 
Section~\ref{sec:adding-exits} describes our approach for incorporating multiple exits into the KAN architecture. 
Section~\ref{sec:results} presents results comparing single-exit and multi-exit KANs across multiple domains.
Section~\ref{sec:learning-to-exit} introduces the learning-to-exit method.
Finally, we conclude in Sec~\ref{sec:discussion} by discussing the implications of our findings, including why multi-exit architectures improve KAN performance, and directions for future work.

\section{Background}
\label{sec:background}

We consider the problems of learning unknown functions from data: nonlinear regression,
\begin{equation}
\mathbf{y} = F(\mathbf{x}) %
\label{eqn:regression-formulation}
\end{equation}
for $\mathbf{y} \in \mathbb{R}^{n\times m}$ and $\mathbf{x} \in \mathbb{R}^{n \times d}$,
and data-driven modeling of continuous or discrete dynamical systems of the form
\begin{equation}
\frac{d \mathbf{x}}{dt} = \mathbf{F}(\mathbf{x})
\label{eqn:dynamical-system-formulation-continuous}
\end{equation}
or
\begin{equation}
\mathbf{x}_{n+1} = \mathbf{F}(\mathbf{x}_n),
\label{eqn:dynamical-system-formulation-discrete}
\end{equation}
respectively, and subject to problem-relevant initial and/or boundary conditions.
KANs have proven effective for both problems~\cite{liu2025kan,PhysRevResearch.7.023037,koenig2024kan}.

\subsection{Kolmogorov--Arnold networks}

A KAN network is a multilayer feedforward neural network but unlike a traditional multilayer perceptron (MLP), the nonlinearities come from learnable, univariate activation functions (Fig.~\ref{fig:overview}) associated with the connections between layers, and fixed summations (or multiplications) propagate signals between layers. 
In contrast, MLPs use fixed nonlinear activation functions on the units and learnable weights for summations associated with the connections between layers.

MLPs are motivated by the universal approximation theorem~\cite{HORNIK1989359} while KANs are motivated by the KART, or \textit{Kolmogorov--Arnold Representation Theorem}~\cite{kolmogorov1961representation,arnold2009functions,kolmogorov1957representations}:
every multivariate continuous function on a finite domain can be expressed as a finite superposition of univariate continuous functions.
More specifically, for $F: [0,1]^d \to \mathbb{R}$, 
\begin{equation}
F(\mathbf{x}) = F(x_1, x_2, \ldots, x_d) = \sum_{q=1}^{2d+1} \Phi_q\left( \sum_{p=1}^d \phi_{q,p}(x_p)\right),
\label{eqn:kart}
\end{equation}
where $\phi_{q,p}:[0,1]\to \mathbb{R}$ and $\Phi_q:\mathbb{R}\to \mathbb{R}$.
The
KART shows that it is possible to represent any continuous function of multiple variables as a composition of one-dimensional functions. 
The innovation of KANs is to operationalize this by learning the univariate ``activation'' functions and stacking them in deep layers. 
As shown by Liu~\emph{et al.}~\cite{liu2025kan}, the stacking, which extends beyond the KART, often allows for smooth and interpretable activation functions which are not expected in the two-layer KART form  given by Eq.~\eqref{eqn:kart}.

In a KAN, the activation functions are typically parameterized using B-splines, local polynomial approximations of the one-dimensional functions. 
Although many other function-fitting techniques have been considered, including %
radial basis functions~\cite{li2024fastkan}, 
Fourier series~\cite{xu2024fourierkan},
sinusoidal functions~\cite{reinhardt2024sinekan},
Chebyshev polynomials~\cite{sidharth2024chebyshev}, and 
wavelet-based representations~\cite{bozorgasl2405wav},
all of which have many advantages, particularly in terms of computational efficiency, for our purposes here we focus on the original B-spline approach, though we also demonstrate the generalizability of our multi-exit architecture using Fourier series in App.~\ref{app:fourier-kan}.

Specifically, each activation function $\phi(x)$ is parameterized as a combination of a base function and a B-spline component:
\begin{equation}
\phi(x) = b(x) + \sum_j c_j B_j(x)
\label{eqn:act-fun}
\end{equation}
where $b(x)$ is the base function, $c_j$ are the learnable coefficients, and $B_j(x)$ are the B-spline basis functions.
The base function serves as a residual connection similar to those in ResNets~\cite{he2016deep}, facilitating gradient flow during training while allowing the B-spline component to learn nonlinear deviations. 
Common choices for the base function include the identity function $b(x) = x$, the SiLU function $b(x)= x/\left(1 + e^{-x}\right)$, or the zero function $b(x) = 0$ when residual connections are omitted.
Additionally, this formulation helps maintain the effectiveness of low-order B-splines even in deep networks by preventing their nested composition from creating numerically unstable high-order polynomials.

To form a deep network by stacking layers of 
learned activation functions, summation units (and, optionally, multiplication~\cite{liu2024kan2.0}) aggregate the outputs from the previous layer's activation functions.
The number of units across $L$ layers is $[d = N_0, N_1, \ldots, N_L=m]$. 
The shape of the KAN is the vector of widths $[N_0, N_1, \ldots, N_L]$.
(See Fig.~\ref{fig:overview} for an example of a KAN network with shape $[2,3,2,1]$.)
Between layers $i$ and $i+1$ there are $N_i N_{i+1}$ activation functions.
Following~\cite{liu2025kan}, $\phi_{\ell,j,i}(x)$ denotes the activation function connecting unit $i$ in layer $\ell$ to unit $j$ in layer $\ell+1$ 
($\ell = 0, \ldots, L-1$; $i = 1, \ldots, N_\ell$; $j = 1, \ldots, N_{\ell+1}$).
The summation units then combine the activation functions to propagate information through the network: for the signal $x_{\ell+1,j}$ into unit $j$ in layer $\ell+1$, giving 
\begin{equation}
x_{\ell+1,j} = \sum_{i=1}^{N_\ell} \phi_{\ell,j,i}(x_{\ell,i}), \quad j = 1, \ldots, N_{\ell+1},
\end{equation}
or, in matrix (broadcast) form, $\mathbf{x}_{\ell+1} = \Phi_{\ell}\left(\mathbf{x}_\ell\right)$, where $\Phi_\ell$ is the functional matrix containing the $\phi$'s connecting layer $\ell$ and $\ell + 1$.
Finally, the full network is represented by composing each $\Phi$ in sequence,
\begin{equation}
\mathrm{KAN}(\mathbf{x}) = \left(\Phi_{L-1} \circ \Phi_{L-2} \circ \cdots \circ \Phi_1 \circ \Phi_0\right) \left(\mathbf{x}\right).
\end{equation}
It is this combination of superpositions of learned activation functions and layer-wise composition that gives KANs both their expressiveness and makes them distinct from MLPs.
Beyond this architectural difference, KANs are considered more interpretable than MLPs because each activation function $\phi$ can be individually examined, as shown in Fig.~\ref{fig:overview}.

KANs are trained with a loss function $\mathcal{L} = \mathcal{L}_\text{data} + \lambda \mathcal{L}_\text{reg}$ comprising a data loss and a regularization loss, with the hyperparameter $\lambda$ determining regularization strength, with $\lambda = 0$ corresponding to no explicit regularization.
Usually the data loss is mean squared error (MSE):
\begin{equation}
\mathcal{L}_{\text{data}} = \frac{1}{n} \sum_{i=1}^{n} \left(y_i - \hat{y}_i\right)^2,
\label{eqn:data-loss}
\end{equation}
where $y_i$ is the true value for observation $i$ and $\hat{y}_i$ is the KAN's prediction for that observation, while the regularization loss is a combination of an L1 norm and an entropy both defined on the activation functions (for details, see Liu \emph{et al.}~\cite{liu2025kan}).
Activation function parameters are learned via gradient-based optimization to minimize $\mathcal{L}$ on training data.
Training commonly includes a refinement process where the resolution of the B-spline grids is increased; gradually increasing grid resolution during training has been shown to improve KAN accuracy better than using a finer grid from the start~\cite{liu2025kan}. 
This grid refinement process acts as a form of path regularization.
In addition to refinement, Liu \emph{et al.}~\cite{liu2025kan} also introduce pruning of weak links as a post-training regularization step (we discuss the potential of pruning in Sec.~\ref{sec:discussion}).
KANs are typically optimized with quasi-Newton methods, particularly L-BFGS~\cite{liu1989limited}, also employed in this work, though first-order methods such as SGD or Adam~\cite{kingma2014adam} can be used as well.

\subsection{Multi-exit and early exit networks}

Multi-exit networks are a deep learning architecture where some or all hidden layers in the network have an additional branch point called an exit~\cite{lee2015deeply,branchyNet,conditionalDeepLearning}. 
These exits are small subnetworks, often only a single layer, whose output can be used alongside or instead of the main trunk network's output.
The most common application is for deep classifiers~\cite{lee2015deeply,branchyNet,bertLosesPatience2020,xin-etal-2021-berxit}, often for computer vision or language models. 
In a classification task, the network is typically given inputs of varying difficulties, %
For easy inputs, i.e., those far from the decision boundary, the network may be able to classify accurately using only basic features built by the earlier layers. 
But for difficult inputs, the network may need to utilize all the layers to build more complex features in order to make a successful classification~\cite{lee2015deeply}.
By equipping the network with multiple exits, an easy input can reduce inference-time compute by using the output of an early exit only, saving both time and energy when making predictions~\cite{scardapane2020should}.
The promise of efficiency gains motivates the study of early exit networks and \emph{learning-to-exit} algorithms.
This approach goes by various names in the literature, including \emph{deeply supervised} networks, \emph{cascaded learning}, \emph{conditional deep learning}, and \emph{adaptive inference}.
For more on multi-exit and early-exit networks, see Scardapane~\emph{et al.}~\cite{scardapane2020should}, Laskaridis~\emph{et al.}~\cite{10.1145/3469116.3470012} or Rahmath~\emph{et al.}~\cite{10.1145/3698767}.

While multi-exits offer energy-efficient and fast inference, 
for our purposes, they provide a more important benefit: they allow \emph{deep supervision}~\cite{lee2015deeply} of the network during training.
By introducing a loss function that combines the outputs of all exits, training gradients enter directly into the earlier hidden layers. 
With appropriate losses, this can allow for a network that is trained more accurately or more efficiently, or both~\cite{scardapane2020should}.

We argue in this paper that multi-exits are a natural extension of KANs and, unlike alternative forms of deep supervision such as DenseNet-style forward connections~\cite{huang2017densely} (see Discussion), this form of deep supervision is especially appropriate to the interpretability advantages of KANs.

\section{Adding exits to KANs}
\label{sec:adding-exits}

A multi-exit KAN augments a standard (single-exit) KAN of shape $[d=N_0, N_1, \ldots, N_L=m]$ with additional exits as follows.
For each layer $\ell$ of the KAN, beginning from the input and continuing until the second-to-last hidden layer\footnote{The last hidden layer already has an exit, the main trunk output.}, add an exit layer, another KAN network, with shape $[N_\ell, m]$. %
The $m$-dimensional output of these exits can then be used to predict the same output as the main trunk exit.
We illustrate a multi-exit KAN alongside the corresponding single-exit version in Fig.~\ref{fig:overview}.

To train a multi-exit KAN requires a data loss and a regularization loss. 
The regularization loss can remain the same as in standard KANs, but applied to all activation functions across the network including those in the exits.
The data loss, on the other hand, must now accommodate multiple predictions. 
A multi-exit KAN with $K$ exits will now emit $K$ outputs, denoted $\hat{y}^{(0)}, \hat{y}^{(1)}, \ldots, 
\hat{y}^{(K-1)}$, where $\hat{y}^{(0)}$ is the output of the exit connected directly to the KAN input layer and $\hat{y}^{(K-1)}$ is the output of the main trunk.
To train the entire KAN across all exits, enabling deep supervision~\cite{lee2015deeply} of all layers, requires a joint loss function that combines all these outputs. 
A straightforward option that we focus on is a weighted average of the individual exit data losses:
\begin{equation}
\mathcal{L}_{\text{multi}} = \sum_{k=0}^{K-1} w_k \mathcal{L}_k,
\label{eqn:exit-weight-loss}
\end{equation}
where $\mathcal{L}_k = \sum_i\left(y_i - \hat{y}_i^{(k)}\right)^2/n$ is the MSE for exit $k$
and $w_k$ is the weight for exit $k$.
The exit weights satisfy $\sum_k w_k = 1$.
These weights become a hyperparameter to be tuned by the researcher using validation data and this tuning process in practice has been straightforward. 
Equation~\eqref{eqn:exit-weight-loss} is quite flexible as the exit weights allow us to prioritize certain exits by weighting them more heavily than others, and individual exits can even be disabled by zeroing their weights.
However, for a network with many exits, these $K-1$ degrees of freedom become a large search space. 
Therefore, after our main results, in Sec.~\ref{sec:learning-to-exit} we propose and apply a ``learning-to-exit'' method to automatically learn $w$ alongside the KAN parameters.

\paragraph{Number of parameters}
With more parameters requiring more training time, it is important to understand
how many more parameters are added to a KAN by adding exits.
The number of parameters in a KAN depends on its architecture, which dictates the number of activation functions, and the number of parameters per activation function.
The parameters per activation function depends on the number and order of the spline bases (Eq.~\eqref{eqn:act-fun}), assuming B-splines are used to model the activation functions. 
This is the same for activation functions in the main trunk and in the exits, so we only need to consider the number of activation functions to determine the overhead added to a KAN by adding exits.

The number of activation functions between layers in a standard KAN is the product of the layer widths, so a KAN with shape $[d = N_0, N_1, \ldots, N_L=m]$ will have 
\begin{equation}
    N_\text{act} = d N_1 + N_1 N_2 + \cdots + N_{L-1} m
\end{equation}
activation functions.
A multi-exit KAN of the same shape will have those activation functions plus an additional $m$ activation functions for each additional exit:
\begin{equation}
\begin{split}
    N_\text{act} &= d N_1 + N_1 N_2 + \cdots + N_{L-2} N_{L-1} \\
    &\quad + \left(d + N_1 + \cdots + N_{L-1}\right)m.
\end{split}
\end{equation}
Notice that the sum of layer widths will be smaller than the sum of products of adjacent layer widths, (unless the layers are all one unit), so, unless $m$ is large, the main trunk dominates the number of parameters in the KAN.
Indeed, for the case of uniform layer width, $N_\ell =d$ for all $\ell$, the main trunk will have 
$(L-1)d^2 + dm$ activation functions, including the original exit, and the newly added exits will introduce $(L-1)dm$ activation functions in total.
In this case, the new exits will contribute fewer activation functions than the main branch when $m < d$, typical of regression problems (Eq~\eqref{eqn:regression-formulation}) and a nearly equal number (fewer by $dm$) of activation functions when $m=d$, typical of dynamical systems modeling (Eqs.~\eqref{eqn:dynamical-system-formulation-continuous} or \eqref{eqn:dynamical-system-formulation-discrete}).

Also, note that no one prediction made by the model will use all the activation functions, even if they were all used during training. 
Thus, multi-exits provide their benefits with reasonable, often modest, parameter overhead.
Training time overhead is similarly modest, as detailed in Appendix~\ref{app:methods}.

\section{Results}
\label{sec:results}

Experiments compare single-exit and multi-exit KANs on various regression tasks of known functional forms (Sec.~\ref{subsec:results:regression}; Figs.~\ref{fig:experiment_1d} and \ref{fig:experiment_2d}; Table~\ref{tab:feynman}), 
on multi-step forecasting of dynamical systems (Sec.~\ref{subsec:results:dynamical-systems}; Figs.~\ref{fig:ikeda-ms} and \ref{fig:food-chain}), 
on a model of continual learning (Sec.~\ref{subsec:results:continual-learning}, Fig.~\ref{fig:continual-learning}), 
and on three real-world datasets ((Sec.~\ref{subsec:results:real-world}, Table~\ref{tab:real_world_kan}).
For each task, a manual architecture search identifies good KAN shapes and exit weights, while holding all other hyperparameters fixed. 
Performance is assessed with the root mean squared error ($\rmse$) of its predictions on test data (as well as $R^2$ values for the real-world data).
Regarding the exit weights, Sec.~\ref{sec:learning-to-exit} explores learning the weights automatically with a ``learning-to-exit'' approach.

\subsection{Regression tasks}
\label{subsec:results:regression}

Our first experiment uses the sinc function, %
\begin{equation}
    f(x) = \frac{\sin\left(\pi x\right)}{\pi x}.
    \label{eqn:sinc}
\end{equation}
This one-dimensional function works well as a test because it features both local oscillations and global decay, and approximation methods often struggle due to its sharp spectral cutoffs, making it a simple but challenging benchmark for function approximation.

\begin{figure}[htbp!]
    \centering
    \includegraphics[width=0.45\textwidth]{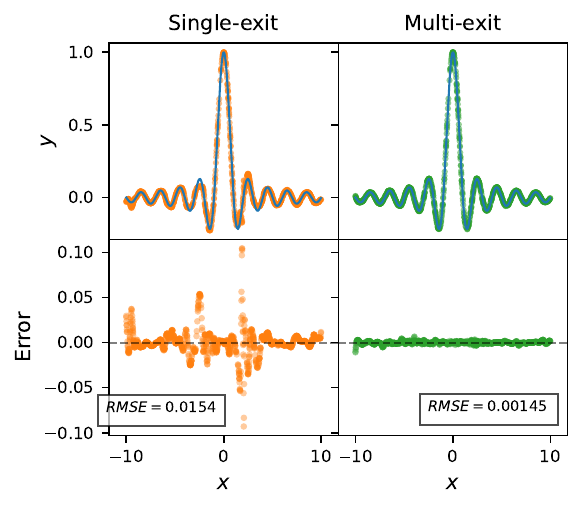}
    \caption{Multi-exits reduce error in 1D nonlinear regression.}
    \label{fig:experiment_1d}
\end{figure}

For the sinc function a KAN of shape $[1, 2, 2, 2, 1]$ performed well.
See Appendix~\ref{app:methods} for full details on training settings and hyperparameters and data generation.
This KAN may seem deep for such a function.
KANs can support multiplication units~\cite{liu2024kan2.0}, although they are not strictly necessary due to the KART, so we decided to forgo them for simplicity and therefore expected a deeper KAN to perform better for this task, hence the aforementioned shape.
Parameterizing the multi-exit weighted loss with a simple linear ramp, $w = [1, 2, 3, 4]$ (unnormalized) worked well: performance on test data was good, with the final exit (shown in Fig.~\ref{fig:experiment_1d}) showing an order of magnitude lower error than the equivalent single-exit network. 
In fact, three of the four exits, all but Exit 0, outperformed the single-exit network, indicating robust and parsimonious (parameter-efficient) capture of the data.

(Note that neither model shown in Fig.~\ref{fig:experiment_1d} is optimal, a point we return to in Sec.~\ref{sec:discussion} as it illustrates important aspects of using a multi-exit architecture.)

\begin{figure*}[htbp!]
    \centering
    \includegraphics[width=1\linewidth]{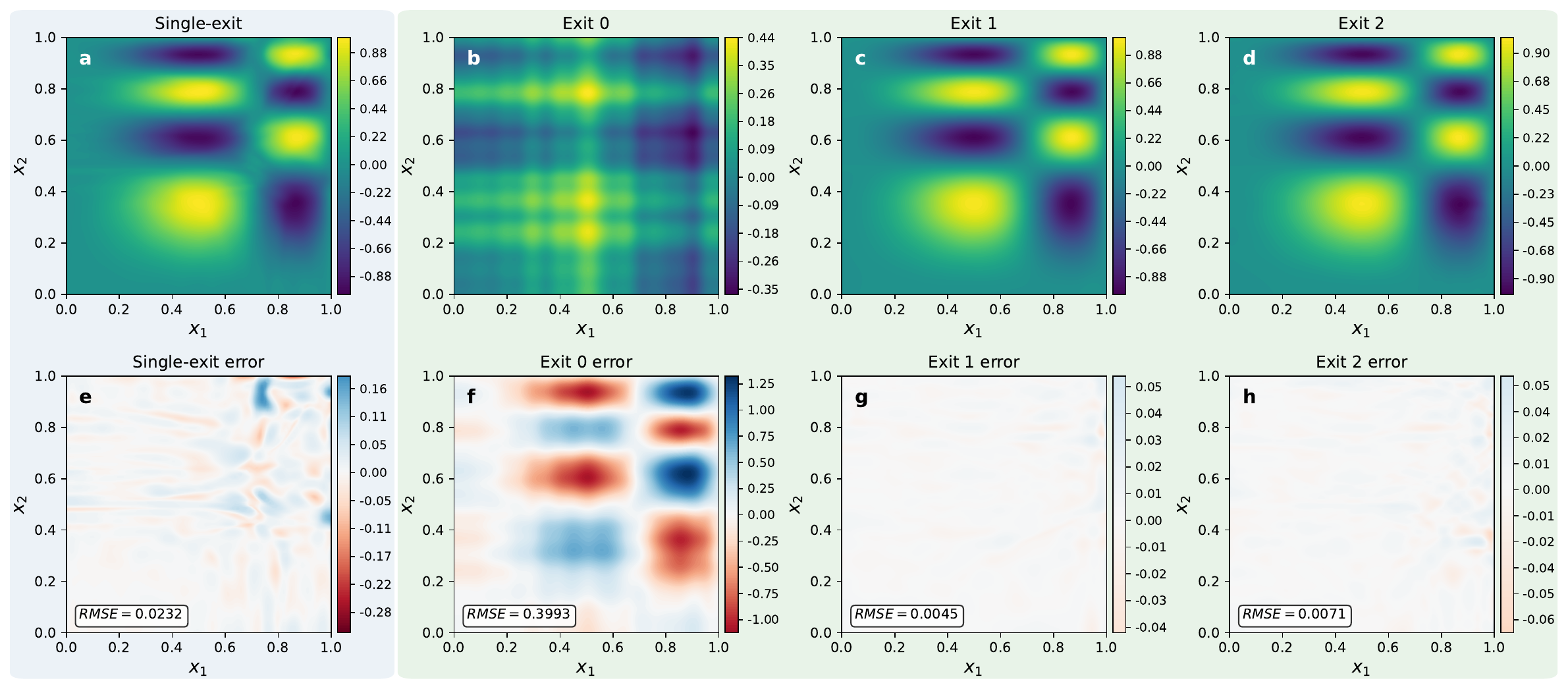}
    \caption{Multi-exits reduce error in 2D nonlinear regression.}
    \label{fig:experiment_2d}
\end{figure*}

Next was the function %
\begin{equation}
    f(x_1, x_2) = \sin\left(2\pi x_1^2\right) \sin\left(4\pi x_2^2\right).
\end{equation}
This nonlinear function tests multivariate approximation through %
spatially-varying frequencies that challenge learning.
Models used a KAN shape of $[2, 3, 2, 1]$ and, for the multi-exit KAN, exit weights $w = [1,2,1]$.
As shown in Fig.~\ref{fig:experiment_2d}, we found good results for the single-exit KAN but even better for the multi-exit KAN.

In the multi-exit KAN,
unsurprisingly, the initial exit, which lacks any composability, is unable to represent this function.
The next exit, however, does well, achieving error one fourth that of the larger, single-exit model ($\rmse =0.0045$ vs.\ $0.0232$). 
The final exit at $\rmse = 0.007$ also outperforms the same-size single-exit model.
This result demonstrates that multi-exit KANs can identify the right level of complexity, with the middle exit outperforming both simpler and more complex alternatives.

Our final regression experiment uses a sample of equations from the \textit{Feynman Equation} dataset, a standard function approximation benchmark~\cite{udrescu2020ai,udrescu2020ai2}.
These equations cover a range of functional forms and complexity levels, while representing practically relevant physical relationships.

Each equation was converted to the dimensionless form indicated in the table and used to generate data (see Appendix~\ref{app:methods}).
KANs with shape $[d, 5, 5, 5, 5, 1]$) were fitted to each dataset.
The multi-exit KAN found good results with $w = [0,0,1,1,3/2]$ (only Exits 2--4 are active) on Eq.~I.27.6, and we proceeded to use this $w$ across all equations.
For each multi-exit KAN, Table~\ref{tab:feynman} reports the smallest $\rmse$ across its exits.

\begin{table*}[th]
    \centering\small
    \begin{tabular}{llllll}
Feynman Eq. & Original Formula & Dimensionless formula & \# vars & $\rmse$ (single) & $\rmse$ (multi) (Exit)\\
\hline
I.6.20 & $\frac{e^{- \frac{\theta^{2}}{2 \sigma^{2}}}}{\sqrt{2\pi\sigma^{2}}}$ & $\frac{e^{- \frac{\theta^{2}}{2 \sigma^{2}}}}{\sqrt{2\pi\sigma^{2}}}$ & 2 & $1.90 \times 10^{-4}$ & $\bm{1.03 \times 10^{-5}}$ (3) \\
I.6.20b & $\frac{e^{- \frac{\left(\theta - \theta_{1}\right)^{2}}{2 \sigma^{2}}}}{\sqrt{2\pi\sigma^{2}}}$ & $\frac{e^{- \frac{\left(\theta - \theta_{1}\right)^{2}}{2 \sigma^{2}}}}{\sqrt{2\pi\sigma^{2}}}$ & 3 & $\bm{1.12 \times 10^{-3}}$ & $1.16 \times 10^{-3}$ (2) \\
I.9.18 & $\frac{G m_{1} m_{2}}{\left(x_{2} - x_{1}\right)^{2} + \left(y_{2} - y_{1}\right)^{2} + \left(z_{2} - z_{1}\right)^{2}}$ & $\frac{a}{\left(b - 1\right)^{2} + \left(c - d\right)^{2} + \left(e - f\right)^{2}}$ & 6 & $5.03 \times 10^{-2}$ & $\bm{2.09 \times 10^{-2}}$ (3) \\
I.12.11 & $q \left(E_{f} + B v \sin {\theta}\right)$ & $1 + a \sin{\theta}$ & 2 & $4.94$ & $\bm{2.41 \times 10^{-2}}$ (4) \\
I.13.12 & $G m_{1} m_{2} \left(1 / r_{2} - 1 / r_{1}\right)$ & $a \left(1 / b - 1\right)$ & 2 & $1.55 \times 10^{-1}$ & $\bm{1.31 \times 10^{-2}}$ (4) \\
I.15.3x & $\frac{x - u t}{\sqrt{1 - \frac{u^{2}}{c^{2}}}}$ & $\frac{1 - a}{\sqrt{1 - b^{2}}}$ & 2 & $1.32 \times 10^{-2}$ & $\bm{2.86 \times 10^{-3}}$ (2) \\
I.16.6 & $\frac{u + v}{1 + \frac{u v}{c^{2}}}$ & $\frac{a + b}{1 + a b}$ & 2 & $2.05 \times 10^{-3}$ & $\bm{3.04 \times 10^{-4}}$ (2) \\
I.18.4 & $\frac{m_1 r_1 + m_2 r_2}{m_1+m_2}$ & $\frac{1 + a b}{1+ a}$ & 2 & $2.53 \times 10^{-4}$ & $\bm{1.90 \times 10^{-4}}$ (4) \\
I.26.2 & $\operatorname{asin}{\left(n \sin {\theta_{2}} \right)}$ & $\operatorname{asin}{\left(n \sin {\theta_{2}} \right)}$ & 2 & $1.22 \times 10^{-3}$ & $\bm{7.62 \times 10^{-4}}$ (4) \\
I.27.6 & $\frac{1}{n / d_{2} + 1 / d_{1}}$ & $\frac{1}{1 + a b}$ & 2 & $1.77 \times 10^{-5}$ & $\bm{1.54 \times 10^{-5}}$ (4) \\
\end{tabular}
    \caption{Performance on Feynman Equation dataset.}
    \label{tab:feynman}
\end{table*}

As shown in Table~\ref{tab:feynman}, multi-exit networks achieved lower test $\rmse$ than the single-exit model in nine of ten cases.
Interestingly, and in line with our observations from the 2D regression shown in Fig.~\ref{fig:experiment_2d}, for half of the problems, the best performing exit is not the final exit, indicating we find more parsimonious (smaller) models with multi-exits that are also more accurate than larger, single-exit models.

\subsection{Data-driven modeling of dynamical systems}
\label{subsec:results:dynamical-systems}

Beyond regression problems, experiments study how multi-exit architectures perform as models of dynamical systems, which present challenging time-series forecasting problems due to their chaotic attractors, evaluating both one-step and multi-step (closed-loop) prediction tasks.

Two dynamical systems are considered.
The first is the \textit{Ikeda map}~\cite{ikeda1979multiple,hammel1985global}, a famous example of a practically motivated discrete-time chaotic dynamical system that does not admit an accurate sparse representation:
\begin{equation}
\begin{gathered}
x_{n+1} = 1 + \mu \left(x_n \cos\left(\phi_n\right)- y_n \sin\left(\phi_n\right)\right),\\
y_{n+1} = \mu\left(x_n \sin(\phi_n) + y_n \cos(\phi_n) \right),
\label{eqn:ikeda}
\end{gathered}
\end{equation}
where
    $\phi_n = 0.4 - 6\left(1 + x_n^2 + y_n^2\right)^{-1}$ 
and bifurcation parameter $\mu = 0.9$.
KANs, unlike sparse regression methods, have been shown to model the Ikeda map well~\cite{PhysRevResearch.7.023037}.

\begin{figure}[t]
    \centering
    \includegraphics[width=0.45\textwidth]{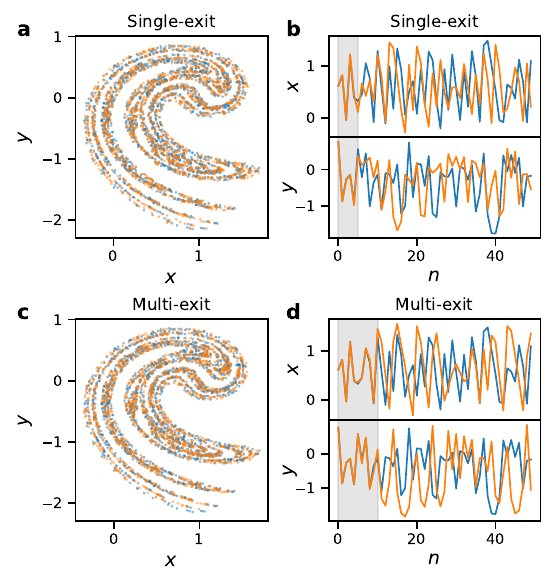}
    \caption{Multi-step prediction of the Ikeda map (Eq.~\eqref{eqn:ikeda}). 
    The multi-exit KAN tracks the dynamics well for about twice as many steps into the future (shaded regions) as the single-exit KAN. (Blue: ground truth; orange: KAN prediction.)}
    \label{fig:ikeda-ms}
\end{figure}

KANs found good results modeling the Ikeda map with a $[2, 4,4,4, 2]$ shape and, for the multi-exit KAN, exit weights $w = [0,0,1,2]$ (the first two exits were disabled).
(Note that Panahi~\emph{et al.}~\cite{PhysRevResearch.7.023037} used a fixed grid $G=10$ while we used grid refinement (see Appendix) for both single- and multi-exit KANs.)
For one-step prediction the multi-exit KAN achieved $\rmse = 4.560 \times 10^{-3}$ compared to the single-exit's $5.484 \times 10^{-3}$, an improvement of $16.8 \%$.
For multi-step prediction, as shown in Fig.~\ref{fig:ikeda-ms} the multi-exit KAN tracks the dynamics for approximately twice as many timesteps as the single-exit KAN before inevitably diverging due to chaos.

The second dynamical system we consider is a continuous-time model of a three-population \textit{ecosystem}:
\begin{equation}
\begin{gathered}
\frac{dN}{dt} = N\left(1 - \frac{N}{K}\right) - x_p y_p \frac{NP}{N + N_0}, \\
\frac{dP}{dt} = x_p P\left(y_p \frac{N}{N + N_0} - 1\right) - x_q y_q \frac{PQ}{P + P_0}, \\
\frac{dQ}{dt} = x_q Q\left(y_q \frac{P}{P + P_0} - 1\right),
\end{gathered}
\label{eqn:food}
\end{equation}
where $N$, $P$, and $Q$ are the primary producer, herbivore, and carnivore populations, respectively, and the carrying capacity $K$ acts as bifurcation parameter.
To model a chaotic system, we set $K = 0.98$, $x_p = 0.4$, $y_p = 2.009$, $x_q = 0.08$, $y_q = 2.876$, $N_0 = 0.16129$, and $P_0 = 0.5$, ensuring the system exhibits a chaotic attractor~\cite{mccann1994nonlinear}.
As with the Ikeda map, KANs are known to model this system well~\cite{PhysRevResearch.7.023037}.

\begin{figure}[t]
    \centering
    \includegraphics[width=0.42\textwidth]{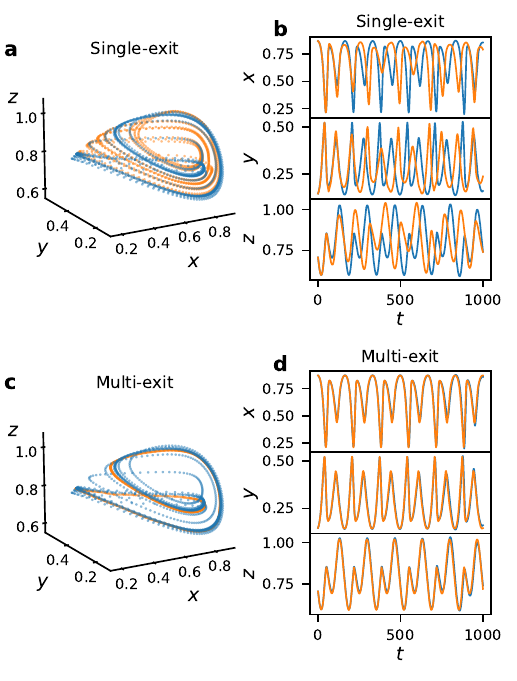}
    \caption{Multi-step prediction of the ecosystem (Eq.~\eqref{eqn:food}).
    The multi-exit KAN encodes the dynamics well, not diverging significantly until $t > 1000$.}
    \label{fig:food-chain}
\end{figure}

KANs performed well modeling the ecosystem with shape $[3, 3, 3, 3]$ KANs and $w= [2, 1, 1/2]$ for the multi-exit KAN, achieving very good multi-step prediction (Fig.~\ref{fig:food-chain}) but slightly worse one-step prediction ($\rmse = 3.774 \times 10^{-4}$ for the multi-exit compared to $3.171 \times 10^{-4}$ for the single-exit KAN).
The single-exit KAN, while still performing well, appears to be overfit to the local trajectory while the multi-exit KAN better captured the underlying attractor structure.

\subsection{Continual learning}
\label{subsec:results:continual-learning}

\begin{figure*}[th]
    \centering
    \includegraphics[width=0.95\textwidth]{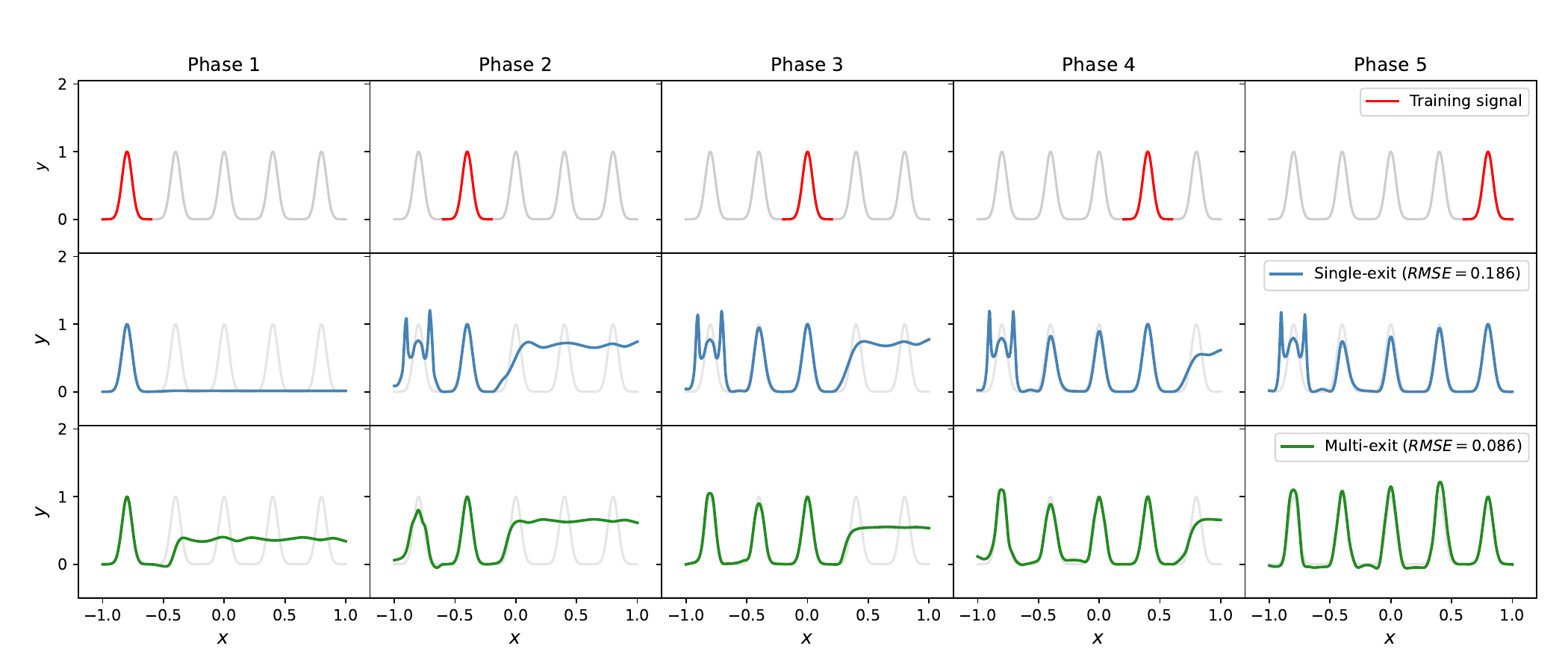}
    \caption{Mitigating catastrophic forgetting with multi-exits in a toy model (Eq.~\eqref{eqn:continual-learning}) of continual learning.}
    \label{fig:continual-learning}
\end{figure*}

As a further demonstration of the usefulness of augmenting KANs with multi-exits, we consider a toy model of continual learning~\cite{liu2025kan,van2024distal} (Fig.~\ref{fig:continual-learning}, top row).
Here the function to represent is a one-dimensional line of five peaks, a multi-modal mixture of Gaussian functions:
\begin{equation}
    f(x) = \sum_{i=1}^{5} \exp\left(-300(x-c_i)^2\right),
    \label{eqn:continual-learning}
\end{equation}
where the centers $c_i$ of peaks $i$ were evenly spaced. 
For this experiment, we generated 100 equally spaced points around each peak, for a total of 500 samples.
KANs can easily fit such data but there is a wrinkle: the model is not trained on all the data at once. 
Instead, it sees the data in phases, one peak at a time (Fig.~\ref{fig:continual-learning}, top row). 

The question becomes whether a KAN can learn a new peak without losing its representation of previous peaks.
As argued by Liu \emph{et al.}~\cite{liu2025kan}, shallow KANs are well adapted to this task due to the local nature of their spline-based activation functions: updating one region of a spline will not affect the fit of other regions, enabling the KAN to retain the form of a previous peak when incorporating the next peak. 
However, they also note that KANs lose this ability as they get deeper, since the composition of splines across multiple layers reduces their locality, opening the door for catastrophic forgetting.
Do multi-exit KANs with their deep supervision retain more locality and exhibit less forgetting?

As seen in Fig.~\ref{fig:continual-learning}, we can answer in the affirmative.
While both architectures display some forgetting, with previously learned peaks changing with subsequent data, the effect is worse for the single-exit KAN (shape $[1,5,5,1]$), especially when learning the second peak.
Compared to the single-exit KAN of the same architecture (Fig~\ref{fig:continual-learning}, middle row), the multi-exit KAN (bottom row) with exit weights $w = [1, 1, 2\text{e}3]$ better tracks the underlying function across training phases, and when finished displays less than half the error of the single-exit KAN ($\rmse = 0.086$ vs.\ $0.19$).

\subsection{Real-world data}
\label{subsec:results:real-world}

Now we consider how multi-exit KANs perform on three real-world datasets:

\begin{itemize}

\item \textit{Airfoil Noise}.
Predict the self-noise (scaled sound pressure, in dB) of a NACA 0012 airfoil for different angles of attack, free stream velocity, and other features. Data originated from anechoic wind tunnel experiments~\cite{brooks1989airfoil}.

\item \textit{Power Plant Energy}.
Predict the electrical power output (in MW) for different ambient atmospheric conditions, temperature, pressure, relative humidity, and the steam turbine pressure (or vacuum).
Data originated from a 480 MW combined cycle power plant with two gas turbines, one steam turbine, and two heat recovery steam generators, and were collected over a six-year period (2006-2011)~\cite{kaya2012local,tufekci2014prediction}.

\item \textit{Superconductor Critical Temperature}.
Predict critical temperature (in K) of superconductors based on material properties.
The dataset contains many features and statistical variants, so for ease of experimentation, we selected five representative features capturing composition complexity, electronic structure, and chemical bonding properties: number of elements, weighted mean valence, valence entropy, weighted mean first ionization energy, and mean electron affinity.
These data originated from the Superconducting
Material Database maintained by Japan’s National Institute for Materials Science~\cite{hamidieh2018data,center-a}.

\end{itemize}

All data were retrieved from the UCI Machine Learning Repository~\cite{kelly_uci} (accessed: 23 May 2025).
From each dataset 1k observations for training and 1k for testing were randomly sampled, except for the smaller Airfoil dataset, which contains only 1503 observations in total so 750 observations for training and 750 for testing were randomly sampled.
Besides sampling for training/testing and selecting features for the superconductivity data, no other filtering or preprocessing was performed.

Single-exit KANs generally performed well with one hidden layer (Table~\ref{tab:real_world_kan}), but we also considered zero- and two-layer KANs, and used the same shapes for the corresponding multi-exit KANs (a KAN with shape $[d,m]$ can only have one exit). 
Experimentation led to exit weights $w$ that performed well, although for both shape and weight, there is likely room for improvement.
All other hyperparameters and training settings (grid refinement, etc.) were unchanged, leaving even more room to improve. 
For multi-exit KANs, in all cases the best performing exit was either Exit 0 or 1.

\begin{table*}[th]
\centering\small
\caption{Performance of single- and multi-exit KANs on three datasets.}
\label{tab:real_world_kan}
\begin{tabular}{lccllccccc}
\toprule
\multirow{2}{*}{Dataset} & \multirow{2}{*}{Obs.} & \multirow{2}{*}{Features} & \multirow{2}{*}{Shape} & \multirow{2}{*}{Exit weight} & \multicolumn{2}{c}{Single} & \multicolumn{3}{c}{Multi} \\
\cmidrule(lr){6-7} \cmidrule(lr){8-10}
& & & & & $\rmse$ & $R^2$ & $\rmse$ & $R^2$ & Exit\\
\midrule
Airfoil           & 1500 & 5         & $[5,1]$     &  ---                     & 5.338  & 0.421   & ---    & ---   & --- \\
                  &      &           & $[5,5,1]$   & $[1, 2\text{e}3]$        & 5.897  & 0.293   & 5.635  & 0.355 & 1   \\
                  &      &           & $[5,5,5,1]$ & $[1\text{e}3, 100, 1]$   & 7.130  & -0.0332 & \textbf{5.129}  & \textbf{0.465} & 0   \\
\midrule
Power plant       & 2000 & 4         & $[4,1]$     & ---                      & 4.462  & 0.934   & ---    & ---   & --- \\
                  &      &           & $[4,4,1]$   & $[5, 2]$               & 4.496  & 0.932   & 4.451  & 0.934 & 0   \\
                  &      &           & $[4,4,4,1]$ & $[100, 1\text{e}3, 1]$   & 4.601  & 0.929   & \textbf{4.406}  & \textbf{0.935} & 1  \\
\midrule
Superconductivity & 2000 & 5 (of 81) & $[5,1]$     & ---                      & 19.524 & 0.658   & ---    & ---   & --- \\
                  &      &           & $[5,5,1]$   & $[10, 8]$                & 18.556 & 0.691   & \textbf{18.353} & \textbf{0.698} & 1   \\
                  &      &           & $[5,5,5,1]$ & $[1, 10, 1]$             & 19.278 & 0.666   & 18.445 & 0.695 & 1  \\
\bottomrule
\end{tabular}
\end{table*}

As seen in Table~\ref{tab:real_world_kan}, multi-exits improved predictive performance on all three datasets.
Interestingly, for all datasets, multi-exit KANs outperformed at early exits the single-exit KANs of the same size and smaller.
For instance, on Airfoil, the three-exit KAN achieved $\rmse = 5.129$ \textit{at Exit 0}, improving on the single-exit KAN of the same size as that exit ($\rmse = 5.338$) by 3.9\%. 
To assess generalizability of these improvements, we evaluated performance on additional test data for two datasets (omitting Airfoil due to data availability).
A one-sided Wilcoxon signed-rank test compared squared residuals between the best single-exit network and the best multi-exit KAN at its optimal exit (as identified in Table~\ref{tab:real_world_kan}).
In both datasets, the residuals of the single-exit network were significantly larger than those of the multi-exit KAN (Power plant: $W = 15\,968\,151$, $p < 0.001$, $n = 7\,568$; Superconductivity: $W = 25\,745\,299$, $p < 0.01$, $n = 10\,000$).
These results confirm that the multi-exit architecture provides statistically significant improvements in predictive accuracy that generalize to new unseen data.

Multi-exit KANs achieved accurate, generalizable fits while simultaneously being more parsimonious on all three datasets.
While additional training of the single-exit KANs could potentially close these gaps, these results nevertheless suggest that multi-exit KANs improve performance simultaneously across shallower and deeper architectures.

\section{Learning to exit}
\label{sec:learning-to-exit}

The exit weight hyperparameter complicates the architecture search problem, already a potential challenge when estimating KAN models.
Indeed, from our experiments, it is not always obvious \textit{a priori} what exit weights will best optimize the KAN. 
Sometimes uniform weights work well, or an increasing or decreasing ramp, or sometimes even a heavy weight on the last exit(s) can be beneficial.
While KANs train quite quickly on these smaller scientific problems, allowing rapid iteration to explore the weight simplex, nevertheless, it would be useful, and data-efficient, to avoid this trial-and-error process.
To this end, here we introduce and apply a basic ``learning-to-exit'' algorithm which treats the exit weights as learnable parameters, eliminating them entirely from the architecture search.

\subsection{Learnable exit weights}

Consider a multi-exit KAN with $L$ exits producing outputs $\{\hat{y}^{(0)}, \hat{y}^{(1)}, \ldots, \hat{y}^{(L-1)}\}$ for a given input. 
Until now multi-exit KANs were trained using a fixed weighted loss (Eq.~\eqref{eqn:exit-weight-loss}):
\begin{equation}
\mathcal{L}_{\text{multi}} = \sum_{i} w_i \mathcal{L}_i\left(\hat{y}^{(i)}, y\right),
\label{eqn:fixed-loss}
\end{equation}
where $w_i$  ($i=0, \ldots, L-1$) are predetermined constants and $\mathcal{L}_i$ is the loss function for exit $i$.
In our learning-to-exit framework, we introduce \emph{exit logits} $\boldsymbol{\theta}_w = \{\theta_{0}, \theta_{1}, \ldots, \theta_{L-1}\}$ to be optimized. 
A softmax transformation connects these to Eq.~\ref{eqn:fixed-loss},
\begin{equation}
w_i(\theta_i) = \frac{\exp(\theta_{i})}{\sum_{j} \exp(\theta_{j})},
\label{eqn:softmax-exits}
\end{equation}
guaranteeing positive, normalized exit weights.
The loss function becomes:
\begin{equation}
\begin{split}
\mathcal{L}_{\text{joint}}(\boldsymbol{\theta}_{\text{KAN}}, \boldsymbol{\theta}_w) = \sum_{i} w_i\left(\theta_i\right) \mathcal{L}_i\left(\hat{y}^{(i)}\left(\boldsymbol{\theta}_{\text{KAN}}^{(i)}\right), y\right),
\end{split}
\end{equation}
where $\boldsymbol{\theta}_{\text{KAN}}$ represents the KAN parameters and $\boldsymbol{\theta}_{\text{KAN}}^{(i)}$ denote the KAN parameters for layers up to and including exit $i$.

\subsection{Optimization procedure}

The optimization problem is formulated as:
\begin{equation}
\boldsymbol{\theta}_{\text{KAN}}^*, \boldsymbol{\theta}_w^* = \arg\min_{\boldsymbol{\theta}_{\text{KAN}}, \boldsymbol{\theta}_w} \mathcal{L}_{\text{joint}}(\boldsymbol{\theta}_{\text{KAN}}, \boldsymbol{\theta}_w).
\end{equation}
Both sets of parameters are updated simultaneously using gradient-based optimization. 
The gradients with respect to the exit logits are
\begin{equation}
\frac{\partial \mathcal{L}_{\text{joint}}}{\partial \theta_{i}} = \sum_{j} \mathcal{L}_j\left(\hat{y}^{(j)}, y\right) \frac{\partial w_j}{\partial \theta_{i}},
\end{equation}
where, from Eq.~\eqref{eqn:softmax-exits},
\begin{equation}
\frac{\partial w_j}{\partial \theta_{i}} = \begin{cases}
w_i(1 - w_i) & \text{if } i = j, \\
-w_i w_j & \text{if } i \neq j.
\end{cases}
\end{equation}

As before, L-BFGS was used for the joint optimization, but 
other methods, such as Adam, could be used instead.
Weight logits were initialized with uniform values ($\boldsymbol{\theta}_w = \boldsymbol{0}$), but other initializations may be worth exploring.
All other training settings and hyperparameters were unchanged.

\subsection{Results}

To evaluate the learning-to-exit model, we first apply it to the three datasets studied in Sec.~\ref{subsec:results:real-world}, focusing on the three-exit architectures.
Comparing to Table~\ref{tab:real_world_kan}, the results were promising.
On every dataset, the learned model outperformed the single-exit KAN of the same shape.
Further, in two of the three cases, the new model outperformed \emph{every} model in Table~\ref{tab:real_world_kan}.
Specifically, for Airfoil the new model achieved $\rmse = 4.947$ and for Superconductivity $\rmse = 18.126$, beating the previous best $\rmse =5.129$ and $18.353$, respectively.
On the other hand, for Power plant, it achieved $\rmse = 4.558$, better than the single-exit result of $\rmse = 4.601$ but worse than the multi-exit's $\rmse = 4.406$.

Interestingly, the learned exit weights varied for all three datasets, despite all being initialized to $w=[1, 1, 1]/3$.
For Airfoil, the final weights were, to machine precision, $w = [1, 0, 0]$ (focus on first exit), for Superconductivity, $w =[0, 0, 1]$ (focus on last exit), and for Power plant, $w =[0.002, 0.848, 0.150]$ (mixed focus).
Power's exit logits also converged more slowly than the other two, which may relate to the weaker relative performance.

Next, we consider the method on some of the earlier synthetic nonlinear regression tasks.
For the sinc function (Fig.~\ref{fig:experiment_1d}), the learning-to-exit algorithm with uniform initial weight converged to weights $w = [0.001, 0.003, 0.761, 0.236]$, favoring Exit 2. 
This configuration achieved $\rmse = 0.00170$, an 89\% improvement over the single-exit baseline ($\rmse = 0.0154$) although not outperforming the fixed weight result of $\rmse = 0.00145$ (which used the deeper Exit 3).
For the 2D regression (Fig.~\ref{fig:experiment_2d}), using a decreasing weight initialization (logits $\boldsymbol{\theta}_w = [1, 0, -1]$), the algorithm converged to weights $w \approx [0, 1, 0]$, focusing entirely on Exit 1 with $\rmse = 0.00756$, a 67\% improvement over the single-exit baseline ($\rmse = 0.0232$) though slightly worse than our previous best results of $\rmse = 0.0045$ and $0.0071$.
A decreasing weight initialization outperformed a uniform one for this task, biasing toward earlier exits and promoting parsimony, suggesting that initialization strategies merit further investigation.

Finally, for the dynamical systems, results were mixed. 
On the food chain ecosystem, the learning-to-exit algorithm with increasing weight initialization converged to $w = [0.108, 0.113, 0.779]$ but achieved $\rmse = 5.18 \times 10^{-4}$, underperforming both the single-exit baseline ($\rmse = 3.28 \times 10^{-4}$) and the previous fixed-weight result ($\rmse = 3.77 \times 10^{-4}$).
For the Ikeda map, the algorithm with uniform initialization converged to $w \approx [0, 1, 0, 0]$, focusing on Exit 1 with $\rmse = 4.66 \times 10^{-3}$, not matching the previous fixed-weight performance ($\rmse = 4.56 \times 10^{-3}$) but showing a modest improvement over the single-exit baseline ($\rmse = 4.82 \times 10^{-3}$).

Overall, our results highlight the learning-to-exit approach while revealing that performance depends critically on initialization strategies and problem characteristics. 
The dynamics of exit selection warrant further study---for one, we expected a regularization term on the exit logits would be necessary, but these results suggest otherwise---yet our findings already show that the learning-to-exit model has promise.

\section{Discussion}
\label{sec:discussion}

Augmenting KANs with multiple exits improves their performance and often their parsimony. 
When a multi-exit KAN performs well at an early exit, it suggests that the full network is deeper than necessary and functions with fewer levels of composition are sufficient to model the given data. 
While in principle a deeper single-exit KAN can be encouraged to simplify, either through regularization or through linearizing the later activation functions---in fact, both single-exit and multi-exit methods are likely to benefit in general from fine-tuning hyperparameters and training settings---nevertheless the multi-exit approach is a promising 
alternative to achieving this parsimony.

What is the mechanism of action behind the success of multi-exits?
The first and more obvious mechanism is that of deep supervision.
The compositional nature of learned activation functions makes gradient flow through many layers a challenge during training.
By connecting the loss function directly to the earlier layers, training will allow for better conditioning of the activation functions and weights within those layers.
In this sense, the multi-exits fulfill a role similar to that of DenseNet~\cite{huang2017densely}-style forward connections. 
In DenseNet architectures, the forward connections link input and hidden layers directly to the final output layer, and backpropagation can then reach deeper into the network for training. 
(The other common form of deep supervision, residual connections (ResNet)~\cite{he2016deep} is already commonly used in KANs; see Eq.~\eqref{eqn:act-fun}.)
In fact, forward connections in KANs could be viewed as another useful form of deep supervision. 
However, they create a large number of outputs at the final layer, which may necessitate a more complex final functional form, hindering KAN's goal of interpretability.
Multi-exits, in contrast, may be a better alternative thanks to their potential for parsimony.

A second and less obvious mechanism of action is implicit regularization through the optimization method. 
In quasi-Newton methods such as BFGS and L-BFGS, a Hessian approximation captures curvature information across \textit{all} parameters simultaneously, enabling the optimizer to find parameter configurations that balance competing objectives from different exits.
In other words, the curvature approximation regularizes the parameters during training.
In our experiments using L-BFGS, the presence of Exit 0 only, the exit connected directly to the input, still led to improvements in KAN performance. 
When there are no other intermediary exits, deep supervision can't condition the earlier layers---Exit 0 is not on the computational path of the last exit. 
Yet, through the optimization process, the network is still able to find better solutions. 
For instance, in the experiments with real world data (Sec.~\ref{subsec:results:real-world}), multi-exit KANs with only one hidden layer still improved, albeit quite modestly, over single-exit KANs. 
Implicit regularization was absent when training with first-order methods such as SGD or Adam~\cite{kingma2014adam}, which update parameters based on individual gradient moments rather than capturing the joint parameter dependencies, though such methods still provide the benefits of deep supervision.
This second mechanism, while weaker than deep supervision, offers another reason why L-BFGS is well-suited for KAN optimization. %

Multi-exit models outperformed all same size and smaller KANs on the real world data (Table~\ref{tab:real_world_kan}), yet we may encounter cases where the best model comes from directly training the appropriate size KAN without using multi-exits.
Indeed, we know this happens, for instance in the sinc function results (Fig.~\ref{fig:experiment_1d}).
A single-exit KAN with one fewer layer than we presented achieved excellent performance ($\rmse = 1.91\times$10$^{-4}$), suggesting that the exit weights shown in the figure were suboptimal.
This observation reveals a key insight: every shallower single-exit network (with otherwise the same layer widths) represents a special case of the multi-exit architecture, corresponding to one-hot exit weights $w = [1, 0, \ldots], [0, 1, 0, \ldots], \ldots, [0, \ldots, 0, 1]$---the vertices of the exit weight simplex.
While these vertices may represent good solutions, \emph{multi-exit architectures can relax into the interior of the simplex} to find even better performance.
Indeed, optimizing the sinc function's exit weights to $w = [2.5\times10^{-4}, 2.2\times10^{-2}, 1\times10^{4}, 0]$ (unnormalized),  an interior solution near the vertex corresponding to the better-performing shallower network, improved performance to $\rmse = 1.80\times10^{-4}$ at the third exit---a relative improvement of over 5\%.
This example demonstrates that multi-exit architectures provide both architecture search capabilities and the potential to discover superior solutions through continuous weight optimization.

While multi-exit KANs introduce additional complexity through exit weights, this does not compromise KAN's core interpretability advantages. 
The activation functions remain visualizable and amenable to symbolic regression~\cite{liu2024kan2.0}, while learned exit weights can themselves provide interpretable insights into layer contributions during training.
Most importantly, multi-exit KANs improve interpretability by identifying more parsimonious configurations---when performance is achieved at early exits, fewer layers of functional composition are needed, yielding inherently more interpretable models.

Despite the promising results presented in this work, some limitations should be acknowledged. 
Perhaps the most serious concern is the extra need to set the exit weights when fitting a multi-exit KAN. 
In all our experiments, setting $w$ required only brief coarse-grained tuning (and it is addressed by our learning-to-exit algorithm) but in settings where data are scarce, it may be difficult to optimize the exit weights without overfitting.
Second, we focused our comparisons on single-exit versus multi-exit KANs to isolate the effect of the architectural change, leaving broader comparisons for future work. 
While comparisons of single- and multi-exit KANs with other interpretable architectures such as decision trees, linear models, or attention-based interpretable networks would provide valuable context, comparing methods with different notions of interpretability calls for different evaluation protocols. Such systematic comparisons across different interpretable architectures are beyond the scope of this work but represent an important direction for future research.
Likewise, future work should consider the effects of different hyperparameter values and training settings, including other sources of regularization such as pruning~\cite{liu2025kan}, to ascertain the optimal settings for different applications and whether multi-exits benefit from or remain robust to such techniques.
Pruning, in particular, while non-differentiable, may interplay with the different exit layers in interesting and useful ways.
Third, while multi-exit architectures often identify more parsimonious models, the interpretability gains are indirect---the exits themselves do not enhance the interpretability of individual activation functions, but rather help identify simpler network configurations. 
Fourth, our learning-to-exit algorithm should be studied further and, while effective, can surely be improved. 
Finally, the additional computational overhead during training, though modest, may become more significant for very deep networks with many exits.

Several promising directions are worth pursuing. 
First, receiving multiple predictions from a KAN immediately brings to mind the idea of ensemble learning~\cite{sagi2018ensemble}. 
However, ensembles benefit from uncorrelated or de-correlated models, but the different exits in a multi-exit KANs are not independent.
Investigating methods to encourage heterogeneity among exits while maintaining their collaborative training could enable ensemble learning.
This may also lead to uncertainty quantification capabilities~\cite{doi:10.1137/1.9781611973228,abdar2021review}---a key goal for KANs~\cite{hassan2024bayesian,mollaali2025conformalized}---—through the natural variation of predictions across exits.
Second, our learning-to-exit framework leaves room for improvement, and incorporating ideas from differentiable architecture search~\cite{liu2018darts} more generally could benefit KANs.
Third, exploring exit architectures beyond simple single-layer KAN exits may yield better performance, although doing so without harming parsimony may be difficult. 
Fourth, extending multi-exit architectures from B-splines to other KAN variants (Fourier-KANs, Wavelet-KANs, Physics-Informed KANs, etc.) could reveal whether the benefits generalize across different basis functions. (We address this in part in App.~\ref{app:fourier-kan}, but more should be done.) 
Finally, applying multi-exit KANs to larger-scale scientific problems, particularly in domains like climate modeling or molecular dynamics where both accuracy and interpretability are crucial, would provide valuable insights into their practical utility.

\paragraph{Conclusion}
We have introduced multi-exit architectures for Kolmogorov--Arnold Networks, demonstrating that augmenting KANs with additional outputs consistently improves their performance across diverse scientific modeling tasks.
Our experiments revealed that multi-exit KANs often achieve their best performance at earlier exits, indicating that they successfully identify more parsimonious models without sacrificing---and often improving---accuracy. 
While multi-exit architectures introduce additional hyperparameters, our results show the benefits substantially outweigh this added complexity.
This finding is particularly valuable for scientific applications where interpretability is paramount, as simpler models with fewer compositional layers are inherently more interpretable. 
These results suggest that multi-exit architectures represent a natural and effective enhancement to KANs, offering researchers a principled approach to finding the right balance between model complexity and performance. 

\appendix

\section{Materials and Methods}
\label{app:methods}

\paragraph{Implementation and training}
KANs were implemented with the PyKAN library v0.2.8 (\url{https://github.com/KindXiaoming/pykan}), based on PyTorch v2.6.0~\cite{paszke2019pytorch}.
To fit KAN models using training data, the Limited-memory Broyden-Fletcher-Goldfarb-Shanno (L-BFGS) algorithm~\cite{liu1989limited} was used with history size 10, strong Wolfe conditions for line search, and convergence tolerances of $10^{-32}$ for gradient norm, parameter changes, and curvature conditions.
Unless otherwise noted, fitting employed a progressive grid refinement strategy, iteratively refining the spline basis functions through a sequence of increasing grid sizes $G = 3, 5, 10, 20$, with 30 optimization steps performed at each grid resolution. 
Spline regularization was not used ($\lambda = 0$) as we found for our data and training settings that it always lowered performance.
Instead, grid refinement acts as implicit regularization and allows the model to first capture coarse-grained patterns before learning finer details, leading to more stable convergence and better generalization performance~\cite{liu2025kan}.
Unless otherwise noted, all other training settings and hyperparameters were kept at the default values of PyKAN, including the learning rate of 1.0 and the default spline order $K=3$ and grid update schedule (grid points were equally spaced and updated every 10 optimization steps).
Multiplication units were not used.
The same training procedure and settings were always used for corresponding single-exit and multi-exit KANs.
Fine-tuning these settings would likely further improve performance, but both types of KANs would be expected to benefit.
Training times were modest with these hyperparameters and data, generally taking 30--90 s for single-exit KANs on a MacBook Pro (M1 Max CPU); the one exception being training on the Ikeda map that took approximately 7 min due in part to the dataset’s high sampling rate.
Multi-exit KANs generally require 50--80\% more time to train than the corresponding single-exit KAN, in line with our parameter count estimates (Sec.~\ref{sec:adding-exits}).
Our source code is available at \url{https://github.com/bagrow/multi-exit-KAN}.

\paragraph{Experimental details}
For the 1D and 2D regression tasks, $n = 1000$ training and $n = 1000$ testing points were generated, with $\mathbf{x} \sim U\left([x_{\min},x_{\max}]^d\right)$ and $y$ generated from $\mathbf{x}$, without additional noise, according to the given equation.
For the fits illustrated in
Fig.~\ref{fig:overview}, $n = 1000$ training points and $n = 200$ testing points were used, as well as a single $G = 5$ grid, 30 optimization steps, KAN shape $[2, 3,2,1]$ and $w = [1,1,1]$.
The Feynman Equations dataset used the same data generating process as the 1D and 2D regression tasks, but each exogenous variable's range was given by the range of values in the released AI Feynman dataset~\cite{udrescu2020ai}.
For the dynamical systems experiments, data for each system was generated and split into training and testing folds following the procedure and parameters of Panahi~\emph{et al.}~\cite{PhysRevResearch.7.023037}. 
For training, a learning rate of 0.1 was used for both systems, and for the Ikeda map specifically, grid updates were not used and 50 optimization steps per grid resolution instead of 30 were used.
Other training settings were unchanged from other experiments.
The continual learning experiment used 100 samples per peak, a grid size of 20 without refinement or updates, 10 L-BFGS steps per phase, and all other settings at PyKAN defaults.
Details for the three real-world datasets were covered in Sec.~\ref{subsec:results:real-world}.

\section{Multi-Exit Fourier KANs}
\label{app:fourier-kan}

To demonstrate that the benefits of multi-exits are not specific to the B-spline formulation, we implemented a Fourier-based variant as preliminary validation of multi-exits across different activation function representations.

\paragraph{Fourier KAN formulation}

Following \cite{fourierKAN} (see also \cite{reinhardt2024sinekan}), each activation function $\phi_{\ell,j,i}(x)$ connecting unit $i$ in layer $\ell$ to unit $j$ in layer $\ell+1$ is parameterized using Fourier series:
\begin{equation}
\phi_{\ell,j,i}(x) = \sum_{k=1}^{G} \left[ a_{\ell,j,i,k} \cos(\omega_k x) + b_{\ell,j,i,k} \sin(\omega_k x) \right]
\end{equation}
where $a_{\ell,j,i,k}$ and $b_{\ell,j,i,k}$ are learnable coefficients, $G$ is the number of frequency components (analogous to grid size in B-spline KANs), and $\omega_k = k \omega_0$ are the fundamental frequencies scaled by a factor $\omega_0$ to prevent high-frequency oscillations over the input domain.
Coefficients are initialized from a normal distribution $\mathcal{N}(0, \sigma^2)$ with $\sigma = 1/\left(\sqrt{d} \left(k+1\right)^2\right)$ to bias learning toward lower frequencies.

The layer output aggregates contributions from all input connections:
\begin{equation}
x_{\ell+1,j} = \sum_{i=1}^{N_\ell} \sum_{k=1}^{G} \left[ a_{\ell,j,i,k} \cos(\omega_k x_{\ell,i}) + b_{\ell,j,i,k} \sin(\omega_k x_{\ell,i}) \right] + b_j
\end{equation}
where $b_j$ is a learnable bias term.

Multi-exits are added and trained as before (Sec.~\ref{sec:adding-exits}) with the exception that analogues of grid refinement and grid updates are not used.

\paragraph{Experimental validation}

We tested Fourier KANs on the polynomial $f(x) = x^3 - 2x^2 + x + 1$ over $x \in [0, 3]$, deliberately choosing a non-trigonometric target to avoid favoring the Fourier parameterization. 
This parallels our use of the sinc function (Eq.~\eqref{eqn:sinc}) for B-spline KANs, ensuring neither method receives an unfair advantage.

We employed a shape $[1, 2, 2, 1]$ architecture with $G = 5$ frequency components and $\omega_0 = \pi/10$, trained using L-BFGS optimization for 30 steps.

The single-exit Fourier KAN achieved an $\rmse= 0.00230$. 
Multi-exit variants with two exit weight configurations demonstrated the effectiveness of the approach:
\begin{itemize}%
\item $w = [0, 1, 100]$: The final exit (Exit 2) achieved $\rmse = 0.00145$, a 37\% improvement.
\item $w = [0, 100, 1]$: The intermediate exit (Exit 1) achieved $\rmse = 0.000761$, a 67\% improvement.
This configuration also outperformed a shape [1,2,1] (same as Exit 1) single-exit KAN that achieved $\rmse = 0.00113$.
\end{itemize}
Notably, the second configuration shows that multi-exit training can identify more parsimonious solutions: the intermediate exit achieved superior performance compared to both the single-exit baseline and the deeper final exit, indicating that a shallower network architecture was sufficient for this task.

These results confirm that multi-exit architectures generalize beyond B-spline parameterizations, improving performance and identifying optimal depths across multiple KAN variants.

{\small
\bibliographystyle{ieeetr}
\bibliography{main}
}

\end{document}